\def\year{2020}\relax
\documentclass[letterpaper]{article} 
\usepackage{aaai20}  
\usepackage{times}  
\usepackage{helvet} 
\usepackage{courier}  
\usepackage[hyphens]{url}  
\usepackage{graphicx} 
\usepackage{graphicx}  

\usepackage{amssymb} 
\usepackage{amsmath} 
\usepackage{diagbox}  
\usepackage{multirow} 
\usepackage{makecell} 
\usepackage{booktabs} 
\title{MIMAMO Net: Integrating Micro- and Macro-motion for Video Emotion Recognition}
\author{Didan Deng,\textsuperscript{\rm 1} Zhaokang Chen,\textsuperscript{\rm 1} Yuqian Zhou,\textsuperscript{\rm 2} Bertram Shi\textsuperscript{\rm 1} \\  
\textsuperscript{\rm 1}Neuromorphic Interactive System Laboratory, Department of Electronic and Computer Engineering, \\The Hong Kong University of Science and Technology, Kowloon, HK\\ 
\textsuperscript{\rm 2} Image Formation and Processing Group, University of Illinois at Urbana-Champaign , Champaign, IL, USA\\
\{ddeng,zchenbc,eebert\}@ust.hk, yuqian2@illinois.edu\\
}
 
\begin{document}
\maketitle
\begin{abstract}
Spatial-temporal feature learning is of vital importance for video emotion recognition. Previous deep network structures often focused on macro-motion which extends over long time scales, e.g., on the order of seconds. We believe integrating structures capturing information about both micro- and macro-motion will benefit emotion prediction, because human perceive both micro- and macro-expressions.
In this paper, we propose to combine micro- and macro-motion features to improve video emotion recognition with a two-stream recurrent network, named MIMAMO (Micro-Macro-Motion) Net. Specifically, smaller and shorter micro-motions are analyzed by a two-stream network, while larger and more sustained macro-motions can be well captured by a subsequent recurrent network. Assigning specific interpretations to the roles of different parts of the network enables us to make choice of parameters based on prior knowledge: choices that turn out to be optimal. One of the important innovations in our model is the use of interframe phase differences rather than optical flow as input to the temporal stream. Compared with the optical flow, phase differences require less computation and are more robust to illumination changes. Our proposed network achieves state of the art performance on two video emotion datasets, the OMG emotion dataset and the Aff-Wild dataset. The most significant gains are for arousal prediction, for which motion information is intuitively more informative. Source code is available at https://github.com/wtomin/MIMAMO-Net.
\end{abstract}

\section{Introduction}
The goal of video emotion recognition is to recognize a subject's emotional state automatically based on videos of their behaviour.
Numerous studies in psychology and neuroscience have proposed ways to quantify human emotions. Seven basic emotion categories were first proposed by ~\cite{ekman1971constants}: anger, fear, disgust, happiness, sadness, surprise, contempt. Even though the seven basic emotion system has become quite popular for multiple emotion recognition systems~\cite{zhou2017action}, people have argued that this system might not be culturally universal~\cite{jack2012facial}. Alternatively, a continuous, dimensional Arousal-Valence space was proposed~\cite{gunes2010automatic}. Arousal refers to how active the emotion is. Valence refers to how positive or negative the emotion is. The Arousal-Valence continuous emotion system has been applied to label a range of emotional datasets ~\cite{barros2018omg,zafeiriou2017aff,kossaifi2017afew}.

Micro- and macro- expressions are both important cues to emotional state. According to \cite{shreve2013automatic}, micro-expressions are described as an involuntary facial expressions which often lasts between 1/25th to 1/3rd of a second (roughly 2-10 frames). Macro-expressions are longer facial expressions which typically last from 3/4th of a second to 2 seconds (roughly 24-60 frames).

Although macro-expression labels, e.g., Facial Action Units (FAU) are often available, and can be recognized automatically~\cite{zhou2017pose}, there are few image datasets labeled with micro-expressions. Nonetheless, we would still like to utilize the information carried by these tiny facial movements. Therefore, we define two broader concepts: micro- and macro-motions. These include both facial movements highly correlated with affective states, i.e., micro- and macro-expressions, and those less correlated, e.g., blinks. Ideally, the model can learn to extract and exploit highly correlated motion features, and to downweight the less correlated motion features.

In existing deep learning models for recognizing spatial-temporal input, there are three common structures: the CNN-RNN ~\cite{fan2016video,yang2018deep,kim2018deep}, the C3D (3D CNN) ~\cite{fan2016video,yang2018deep} and the Two-Stream Network ~\cite{pan2019deep,feng2018dynamic}. The CNN-RNN architecture takes advantage of both the transferred knowledge of a pretrained CNN and the temporal modeling capability of the RNN. The input features to the RNN are usually abstract and global features represented by higher layers. It makes this architecture well suited to extract larger and more sustained changes in facial appearance (i.e., macro-motion).The C3D combines information over both space and time using convolutional filters starting from the lowest layers. This enables it to capture both macro- and micro-motion. However, it cannot incorporate transferred knowledge as conveniently as the CNN-RNN. 

The Two-Stream Network has been less well studied than the CNN-RNN and the C3D. It contains two parallel convolutional networks: a spatial network that processes a static image, and a temporal network that processes instantaneous motion information, most commonly represented by the optical flow. It has the advantages that it can use pretrained CNN features in the spatial stream, and can learn low-level short duration facial movement features in the temporal stream. This makes it well suited for capturing micro-motions. 

In this work, we propose the MIMAMO Net architecture which uses a Two-Stream Network followed by an RNN to capture both macro- and micro-motion efficiently.

One of the most important innovations in this work is the use of phase differences between successive frames of outputs from a complex steerable pyramid to represent low-level motion information. Phase differences can serve as a good alternative to the more common optical flow, primarily because of difficulties in extracting robust and accurate flow vectors. Many optical flow algorithms assume the brightness constancy constraint, which assumes that corresponding points across two frames have the same brightness, as this leads to a simple relationship between the flow vectors and the image spatial and temporal gradients. However, this makes the outputs sensitive to global changes in illumination over time. In addition, since gradients are noisy, significant computational effort must be expended in regularizing the resulting flow estimates, e.g. through the imposition of a smoothness constraint~\cite{brox2004high,horn1981determining}. This makes flow algorithms computationally complex, and can make the algorithms less capable of dealing with non-rigid motion, as expected for facial expressions. In contrast, phase differences of the outputs of a complex steerable pyramid are computationally less demanding, and are invariant to changes in global illumination. 

Our primary contributions are:
\begin{quote}
    \begin{itemize}
        \item We propose a Two-Stream Network followed by an RNN, the MIMAMO Net, to learn features that represent both micro- and macro-motion in the face. Intuitively, micro- and macro-motion are reflective of micro- and macro-expressions. However, we do not explicitly seek to identify these expressions. 
        \item We show that using phase differences, rather than the optical flow, as input to the temporal stream leads to better performance on video emotion recognition. Based on experiments where we randomly varied the brightness from frame to frame, we suggest that the optical flow algorithms are sensitive to illumination changes because of the brightness constancy constraint, whereas phase differences are invariant to changes of the overall illumination.
        \item The proposed MIMAMO Net leads to state of the art performance on the OMG emotion and the Aff-Wild datasets. 
    \end{itemize}
\end{quote}

\section{Related Work}

Prior work applying the Two-Stream Network to video emotion recognition has used the optical flow or Local Binary Patterns from Three Orthogonal Planes (LBP-TOP) as input to the temporal stream. Pan et al.~\cite{pan2019deep} used two CNNs to extract features from RGB images and optical flow images. The extracted features were fed into two separate LSTMs for emotion prediction. Feng et al.~\cite{feng2018dynamic} also utilized two CNNs, where one CNN processed the RGB frame, and the other CNN processed the LBP-TOP features along the x-t and y-t axes. 

We identify several shortcomings in those inputs. First, computing the optical flow or the LBP-TOP composes a heavy computational burden to the system. Second, many optical flow algorithms have the brightness constancy constraint or the spatial smoothing constraint. The former makes optical flow sensitive to illumination changes, while the latter limits the non-rigid motion information in optical flow. 

The systems proposed by~\cite{pan2019deep} and~\cite{feng2018dynamic} were tested only on \textit{in-the-lab} datasets, where head motions were small and illumination conditions changed rarely. Many emerging \textit{in-the-wild} video emotion datasets are developed, including AFEW dataset  ~\cite{dhall2015video},  Aff-Wild datset ~\cite{zafeiriou2017aff} and OMG dataset ~\cite{barros2018omg}. We think that for \textit{in-the-wild} emotion datasets, we should use phase differences, which are more robust to illumination changes than optical flow.

To obtain phase differences, the Complex Steerable Pyramid ~\cite{portilla2000parametric} is often used. It is a multi-scale and multi-orientation image decomposition method, used in a wide variety of image processing and computer vision tasks, such as motion magnification~\cite{wadhwa2016eulerian}. The recent paper ~\cite{duque2018micro} used Riesz Pyramid (a variant of Complex Steerable Pyramid) for micro-expression spotting, which supports our expectation that phase differences will be useful for emotion recognition.

In our proposed method, we integrate a Complex Steerable Pyramid with a Convolutional Neural Network as a cascade in the temporal stream. In terms of using phases or phase differences to detect emotion, our work is most similar to the work of ~\cite{duque2018micro}. They first computed the phases of facial images in the video. Then they divided the phases of the face image into five different areas: two eye areas (left and right eye), and three facial features areas (left and right eyebrow and mouth area). But they simply calculated the variances of the phase signals in those five areas and used peak analysis to spot micro-expressions. In contrast, we use a Two-Stream Network with an RNN to learn more complex spatial-temporal features.

\section{The Complex Steerable Pyramid} \label{sec:complex}
The Complex Steerable Pyramid ~\cite{portilla2000parametric} is a linear image decomposition method. 
It decomposes an image into sets of coefficients corresponding to frequency sub-bands and orientations. From the complex coefficients, we can derive the magnitudes and the phases. 

Phases at each pixel are also called local phases. Local phase shifts are proportional to local displacements. We demonstrate their relationship using Gabor function as a basis function. As shown in Eq.~\ref{eq: gabor}, the Gabor basis function is an oriented complex sinusoid windowed by a Gaussian envelope. $\delta$ is the standard deviation of the Gaussian envelope and $\omega$ is the frequency of sinusoid. The basis function consists of the real part and the imaginary part, which are in approximate phase-quadrature:
\begin{equation}
\label{eq: gabor}
    B(x) = e^{\frac{-x^2}{2\delta ^2}}e^{j\omega x} = e^{\frac{-x^2}{2\delta ^2}} (cos\omega x +j sin\omega x)
\end{equation}
Suppose the input signal is a Dirac delta function $p(x)$ at frame 0 and translated by $x_0$ to $p(x-x_0)$ at frame 1. The filter responses of the basis function in Eq.~\ref{eq: gabor} at frames 0 and 1 are:
\begin{equation}
    e^{\frac{-x^2}{2\delta ^2}}e^{j\omega x},  e^{\frac{-(x-x_0)^2}{2\delta ^2}}e^{j\omega (x-x_0)}
\end{equation}
The phase difference $-\omega x_0$ is proportional to the spatial displacement $x_0$. Therefore, the phase differences represent the spatial displacement in certain frequency band defined by $\omega$. 

The input to the Complex Steerable Pyramid is a two-dimensional image, which outputs a set of coefficients $R_{w,\theta} (x,y)$, where $\omega$ defines the band-pass frequency and $\theta$ defines the orientation angle. Based on $R_{w, \theta}$, we can derive local amplitude and local phase using the following formulas:

\begin{align} 
 R_{w,\theta}(x,y) & =  \operatorname{Re}_{w, \theta}(x,y)+ j\operatorname{Im}_{w, \theta}(x,y) \\
    A_{w,\theta}(x,y) &= \sqrt{\operatorname{Re}_{w, \theta}(x,y)^2 + \operatorname{Im}_{w, \theta}(x,y)^2} \\ 
    \Phi_{w, \theta}(x,y) &= \arctan (\operatorname{Im}_{w, \theta}(x,y)/\operatorname{Re}_{w, \theta}(x,y)) 
\end{align}

Local phase difference cannot be simply obtained by calculating difference between the phases of consecutive frames, because local phase is a wrapped quantity. To unwrap phase, we use the technique from ~\cite{gautama2002phase}, where we add or subtract $k\cdot 2\pi$ if the local phase difference between consecutive frames exceeds $\pi$. Note that this phase-unwrapping method assumes the magnitude of motion between consecutive frame is at most $\frac{\pi}{\omega}$. This limitation can be compensated for by using multiple scales.

\section{Proposed Method}
\subsection{Face Detection and Alignment}
We use the OpenFace Toolkit ~\cite{baltruvsaitis2016openface} to detect faces and 68 facial landmarks from video frames. The facial landmarks are used to define a window to crop the face. The cropped faces are aligned and then resized to 224x224 pixel frames and 48x48 pixel frames, which are inputs to the Two-Stream Network. The face alignment procedure reduces rigid motions of the entire face due to head movements, allowing the Two-Stream Network to focus on extracting information about the facial expression and its changes over time. In the future, it may be interesting to include information about head movements, as they may also contain information related to the emotion state.
\subsection{Phase and Phase Difference Image}
\begin{figure}[ht]
\centering
\includegraphics[width=0.95\columnwidth]{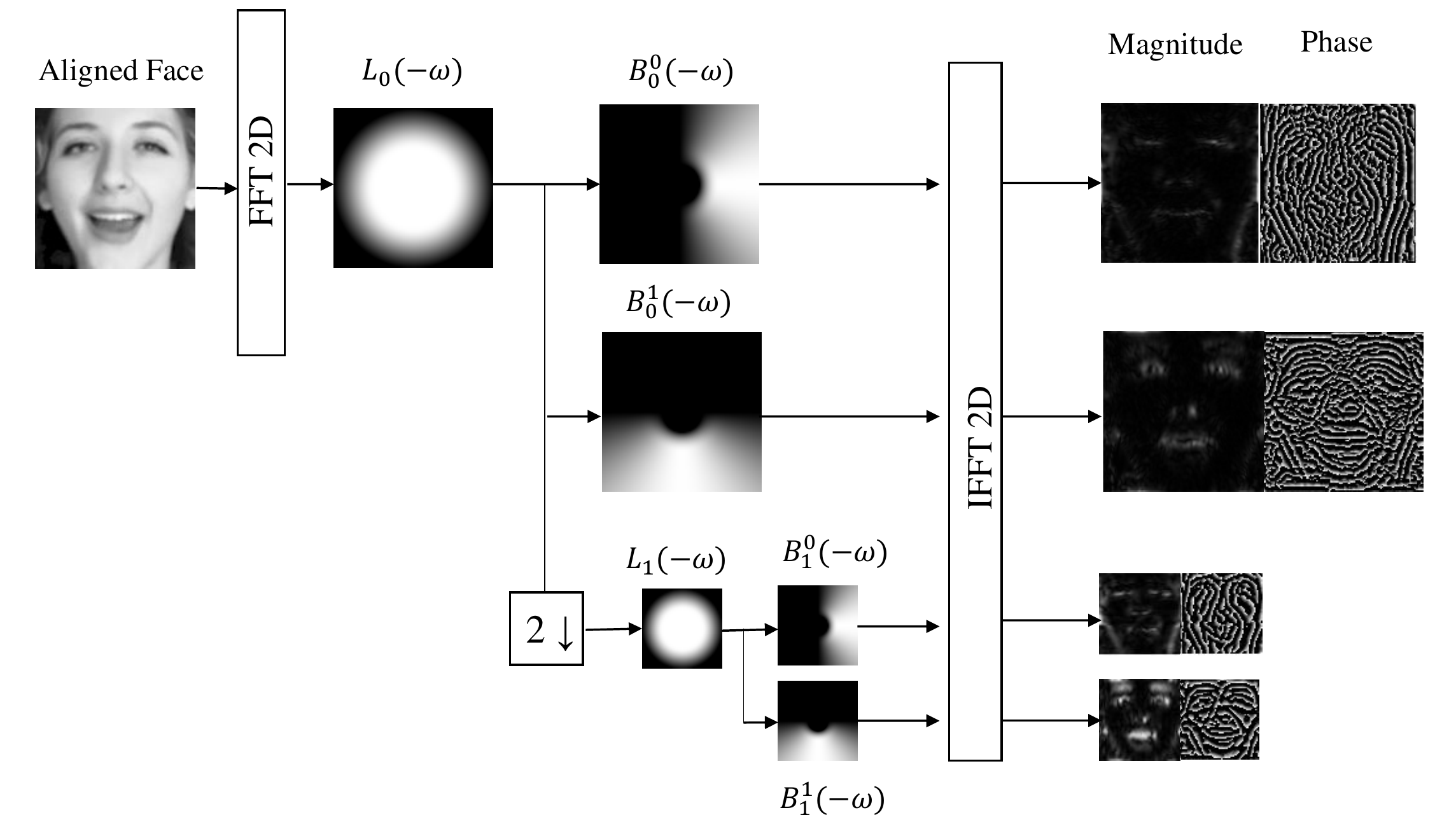} 
\caption{\textbf{Complex Steerable Pyramid}: A diagram for a 2-scale, 2-orientation Complex Steerable Pyramid ~\cite{portilla2000parametric}. The highpass residual and the lowpass residual are omitted here because we do not include them in our method. Low-pass filters and oriented band-pass filters are shown in frequency domain.}
\label{fig: sp}
\end{figure}  

We compute the phase and phase difference images from consecutive aligned facial frames. Fig.~\ref{fig: sp} shows a 2-scale, 2-orientation Complex Steerable Pyramid. The input is a gray-scale facial image. First, we compute the Fast Fourier Transform (FFT) of the input image. Then the FFT is multiplied with the low-pass filter $L_0(-\omega)$ in the first scale. Next, the output of the low-pass filter $L_0(-\omega)$ is multiplied with the frequency response of the band-pass filters in two orientations $B_0^0(-\omega)$ and $B_0^1(-\omega)$. After the inverse FFT on the outputs of band-pass filters, we can obtain the magnitude and phase images of filter responses in two orientations for the first scale of the pyramid. The same operates are applied to a second coarser scale. More orientations and more scales are technically feasible, but to make a fair comparison with optical flow and to reduce computation, we only choose two orientations and two scales here.

\begin{figure}[!ht]
    \centering
    \resizebox{0.7\columnwidth}{!}{
    \includegraphics[width=0.9\columnwidth]{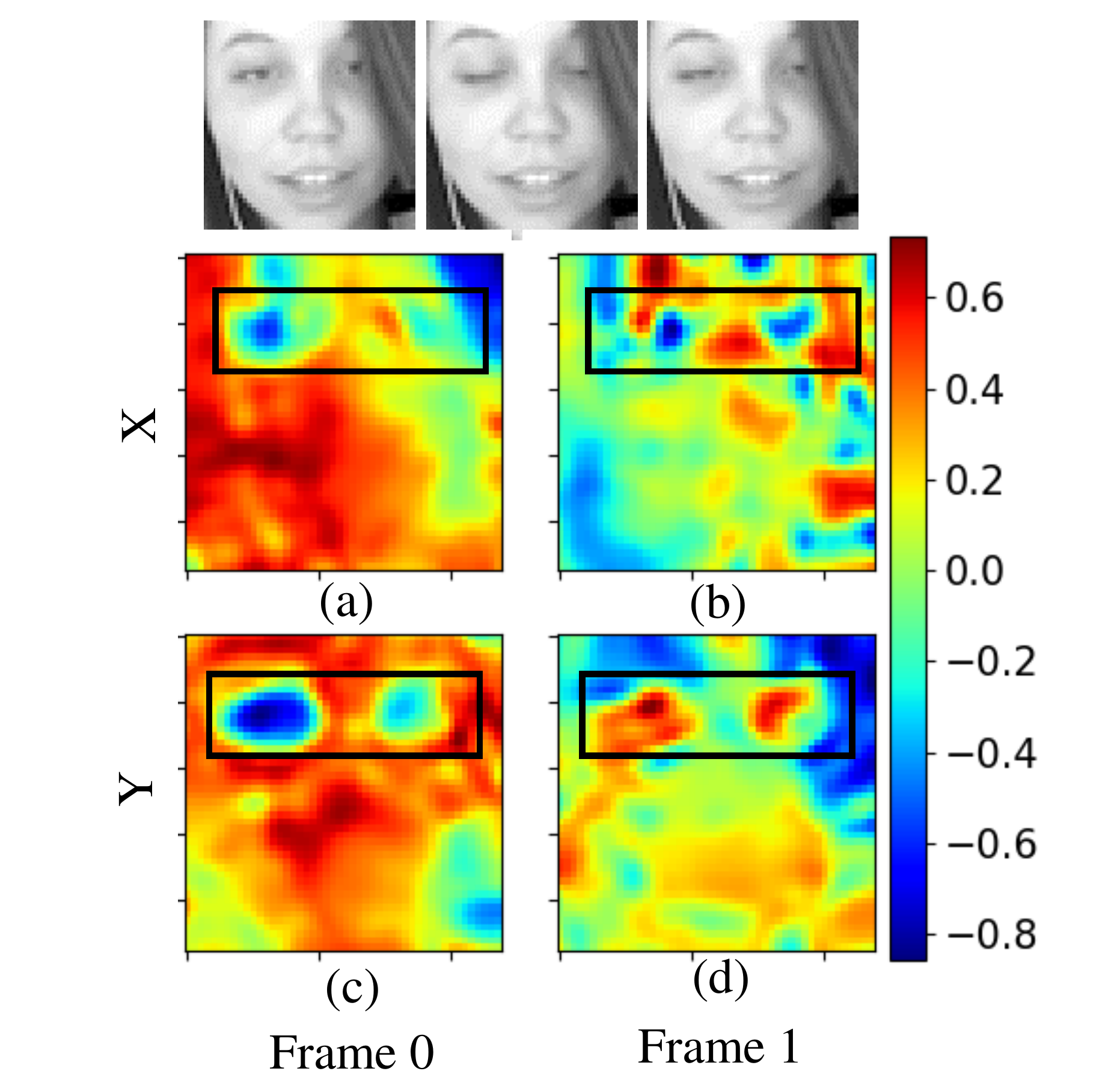} 
    }
\caption{\textbf{Examples of phase difference images}: For the input facial images, we display the phase difference images. Phase difference images have four dimensions: time, orientation, width and height.}
\label{fig: pd}
\end{figure}

With multiple consecutive facial frames as inputs to the Complex Steerable Pyramid, we obtain multiple phase and magnitude images. We first adopt the de-noising method in ~\cite{wadhwa2013phase}, since phase is less reliable where the amplitude is small. We apply an amplitude-weighted spatial Gaussian blur on the phase to improve SNR. Then we unwrap the phase and calculate difference between consecutive phase images to obtain the phase difference images. As proposed by ~\cite{duque2018micro}, the rigid motion between frames can be further reduced by subtracting the mean value of phase differences over a spatial window. We do this to reduce the effect of head movements, which can be modeled locally as translational motion. Fig.~\ref{fig: pd} shows examples of extracted phase difference images in the X direction (a, b) and the Y direction (c, d). We use a smaller input image resolution (48x48) when computing phase differences to match the range of motions represented by the phase differences to the expected range of micro-motions.
\begin{figure*}[ht] 
\centering
\includegraphics[width=\textwidth]{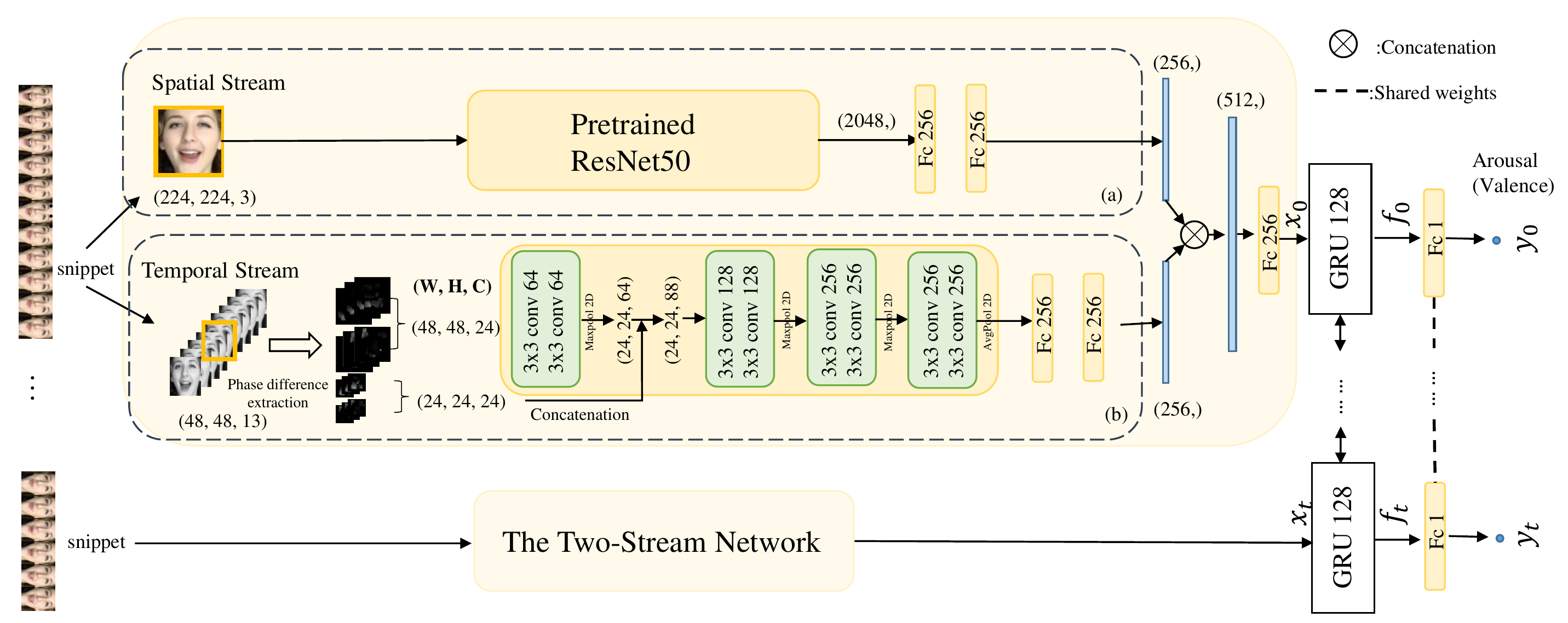} 
\caption{\textbf{MIMAMO Net}: Our proposed model consists of two stages. The first stage is a Two-Stream Convolutional Neural Network. The second stage is a Gated Recurrent Unit network. The figure shows the case when the length of the input to the temporal stream is 13 frames, i.e., the number of phase difference frames is 12. We treat images from different frames and orientations as belonging to different channels (C). (a) and (b) denote the temporal stream and the spatial stream respectively.}
\label{fig: model}
\end{figure*} 
\subsection{MIMAMO Net}
Our proposed model (MIMAMO Net) is illustrated in Fig.~\ref{fig: model}. It consists of two stages. The first stage is a Two-Stream Convolutional Neural Network. The second stage is a Gated Recurrent Unit (GRU) model. The first stage learns a feature representation of a snippet, which consists of an RGB image and a sequence of gray-scale images, centered in time around the RGB image. The RGB image is fed into the spatial stream of the Two-Stream Network. The sequence of gray-scale images are fed into the temporal stream of the Two-Stream Network. The second stage learns the long-term temporal correlations between consecutive snippets.

In the spatial stream, we use the pretrained ResNet50 model~\cite{albanie2018emotion} as feature extractor. The model
is pretrained on the VGGFace2 face recognition dataset~\cite{cao2018vggface2}, and then fine-tuned on the facial expression
static image dataset FER2013 plus~\cite{BarsoumICMI2016}.
We apply average pooling to the final convolutional layer,
resulting in a 2048 dimensional feature vector. This vector is then fed
into two fully connected layers, resulting in a 256 dimensional
final feature vector.
In the temporal stream, after obtaining the sequence of phase difference images, we merge the time and orientation dimensions, resulting in $12\times2$ channels. These are fed into the CNN in the temporal stream, which contains eight convolutional layers divided into four convolutional-layer blocks. Because the coefficients in two scales of the steerable pyramid outputs have different sizes, we concatenate the second-scale inputs with the outputs of the first convolutional block along the channel dimension. 

The outputs of the spatial and temporal stream are concatenated into a 512 dimensional feature vector, which is fed into a fully-connected layer that reduces the dimension to 256.

The output of the Two-Stream Network is fed into a bidirectional GRU network with 128 hidden units. This network combines information across snippets over the entire video sequence, resulting in a 128 dimensional feature vector $f_t$ which is then fed into a fully-connected network outputting the estimated arousal and valence at time $t$. When the frame-level labels are available, we use the outputs of the final linear output layer as predictions. When only video-level labels are available, we average the feature vectors to get $\Bar{f} = \frac{1}{N}\sum_{t=0}^N f_t$, and feed $\Bar{f}$ to the final output layer to produce a video-level prediction.

\subsection{Loss function} We use the Concordance Correlation Coefficient (CCC) as a performance metric. The CCC is defined by:
\begin{equation}
CCC =  \frac{2\rho \sigma_{x} \sigma_{y}}{\sigma_{x}^2 + \sigma_{y}^2 + (\mu_{x} - \mu_{y})^2}
\label{eq:loss}
\end{equation}
where $\rho$ is the Correlation Coefficient, $\mu_{x}$ and $\mu_{y}$ denote the means of the predictions and the ground truth, and $\sigma_{x}^2$ and $\sigma_{y}^2$ are the corresponding variances. The weights of the GRU and the Two-Stream Network are trained by minimizing -CCC.

\section{Implementations}
\subsection{Datasets}

\subsubsection{The OMG-Emotion Behavior Dataset.} The videos in the OMG dataset~\cite{barros2018omg} were collected from YouTube with no environmental constraints. Annotators separated videos into clips based on utterances. Every utterance was assigned with two continuous emotion labels: arousal in [0,1] and valence in [-1, 1]. The number of utterances in the training, validation and test sets are 2441, 617 and 2229 respectively. During our experiments, we merge the training set and the validation set to a larger training set for cross validation.

\subsubsection{Aff-Wild Dataset.} This dataset~\cite{zafeiriou2017aff} contains videos collected from YouTube in real-world settings. There is no overlap between OMG dataset and Aff-Wild dataset. The annotation of Aff-Wild dataset is on the frame-level, where each frame in videos has two labels: arousal in [-1,1] and valence in [-1,1]. In total there are 1,008,650 frames in the training set, and 215,450 frames in the test set. No validation set is provided.

\subsection{Experiments}
\subsubsection{Training.}
During training, we used stochastic gradient descent optimizer implemented in PyTorch, with momentum (0.9) and weight decay ($5e^{-4}$). The number of epochs was set to be 25. Early stopping (5 epochs) was used to prevent overfitting. Batch size varied among [16, 32, 64, 128, 256, 512]. Because different experiments had different GPU consumption, we chose the largest batch size that fits in our GPU memory (11GB). Augmentation methods such as random cropping and horizontal flipping were used. The pretrained ResNet50 model weights were fixed during training, but all the other layers were trainable. Batch normalization and dropout were used after every fully-connected layer.

\subsubsection{Sampling.} When sampling video frames in the OMG dataset, we used a fixed sample rate in which we took one snippet every second. We randomly chose the start of sampled snippets to make sure that we use every frame in training set. But in the test set, we only sampled a video once. When sampling frames in the Aff-Wild dataset, we took the original frame rate as the sample rate because the labels in the Aff-Wild dataset are frame-level. We set the maximum length of input snippets to the GRU model to be 64 due to the limitations in the GPU memory.

    \subsubsection{Training.} We trained two types of networks: single stream and two stream. For the single-stream networks, we excluded the GRU network in Fig.~\ref{fig: model}, and added one linear output layer to the last fully-connected layer with 256 neurons in Fig.~\ref{fig: model} (a) or (b). For video-level prediction, we combined information across all frames by averaging. For the two-stream networks, we used the architecture defined in Fig.~\ref{fig: model}. The weights in Fig.~\ref{fig: model} (a) and (b) were not trained from scratch. Instead they were initialized with the model weights learned in the single-stream networks. 

\subsection{Results on OMG}

Table~\ref{tab:omg_sota} compares the performance of our proposed model with that of the existing methods on the test set of the OMG dataset. Peng et al.~\cite{peng2018deep} used a SphereFace CNN ~\cite{liu2017sphereface} to extract facial features and fed the features into a bidirectional LSTM. A linear output layer after the LSTM gave arousal and valence predictions. The model of Kollias et al.~\cite{kollias2018multi} is similar, consisting of a VGG16 model and a two-layer GRU model. These two methods were single modal methods. Both used CNN-RNN architectures. Their performances were not as good as ours, because we used the temporal stream to learn motion representations. MIMAMO Net outperformed the state-of-the-art single-modal (visual) model~\cite{peng2018deep} by 54.5\% (arousal) and 21.0\% (valence).

Deng et al.~\cite{deng2018multimodal} concatenated the features extracted by the VGG Face model and the features extracted by OpenFace, and then fed the fused features into an LSTM model for the visual modality. Zheng et al.~\cite{zheng2018multimodal} used a pretrained VGG16 model and an LSTM model with an Attention layer for the visual modality. Both methods were multimodal methods. For the audio modality, they both used opensmile~\cite{eyben2010opensmile} to extract hand-crafted audio features. Our model outperformed both bimodal (audio-visual) models, despite the fact that we only used the visual modality. MIMAMO Net's performance was 5.8\% higher for arousal CCC and 6.6\% higher for valence CCC than the best bimodal model~\cite{zheng2018multimodal}. We expect that integrating the audio modality into MIMAMO Net would give even greater gains.

	\begin{table}[b]
	\small
		\begin{center}
		\resizebox{\columnwidth}{!}{
			\begin{tabular}{c|c|c}
				\toprule
				Methods & Arousal CCC & Valence CCC\\
				\cline{1-3}
				 \cite{peng2018deep} & 0.244 & 0.437\\
				\cite{kollias2018multi} & 0.130 & 0.400 \\
				\hline
				 \cite{deng2018multimodal} &\ 0.276 & 0.359\\
				\cite{zheng2018multimodal} &0.356 &0.496 \\
				\cline{1-3}
				MIMAMO Net (Ours)  &\bf 0.377 &\bf 0.529 \\
				\bottomrule
			\end{tabular}
			}
		\end{center}
		\caption{\textbf{Comparison with existing methods} on the test set of the OMG dataset.}
		\label{tab:omg_sota}
	\end{table}

\subsection{Results on Aff-Wild} 
Table~\ref{tab:aff_sota} compares our proposed method with existing methods on the test set of the Aff-Wild dataset. Li et al.~\cite{li2017estimation} used deep convolutional residual neural network for facial feature extraction. And they used multiple memory networks to model the temporal correlation between frames. Their architecture is very similar to the CNN-RNN. Chang et al.~\cite{chang2017fatauva} proposed a deep neural network with an attribute layer, an AU layer (facial action units) and a V-A layer (Valence-Arousal) which were trained sequentially. Zafeiriou et al. ~\cite{zafeiriou2017aff} used a VGG Face model for facial feature extraction, and used a GRU model for temporal modeling. The main difference between MIMAMO Net and their methods is that we employ a Two-Stream Network for spatial-temporal representation learning.

Compared with the SOTA CNN-RNN model~\cite{zafeiriou2017aff}, MIMAMO Net achieved significantly better performance on arousal (21.1\% higher) and slightly better performance on valence (1.7\% higher).

	\begin{table}[b]
	\small
		\begin{center}
		\resizebox{\columnwidth}{!}{
			\begin{tabular}{c|c|c}
				\toprule
				Methods & Arousal CCC & Valence CCC\\
				\cline{1-3}
				 \cite{li2017estimation} & 0.214 & 0.196\\
				 \cite{chang2017fatauva} & 0.282 & 0.396\\
				\cite{zafeiriou2017aff} & 0.430 & 0.570 \\
				\cline{1-3}
				MIMAMO Net (Ours)  & \bf 0.521 & \bf 0.580 \\
				\bottomrule
			\end{tabular}
			}
		\end{center}
		\caption{\textbf{Comparison with existing methods} on the test set of the Aff-Wild dataset.}
		\label{tab:aff_sota}
	\end{table}
	
\begin{figure}[t]
\centering
    \resizebox{0.95\columnwidth}!{
    \includegraphics[width=0.9\columnwidth]{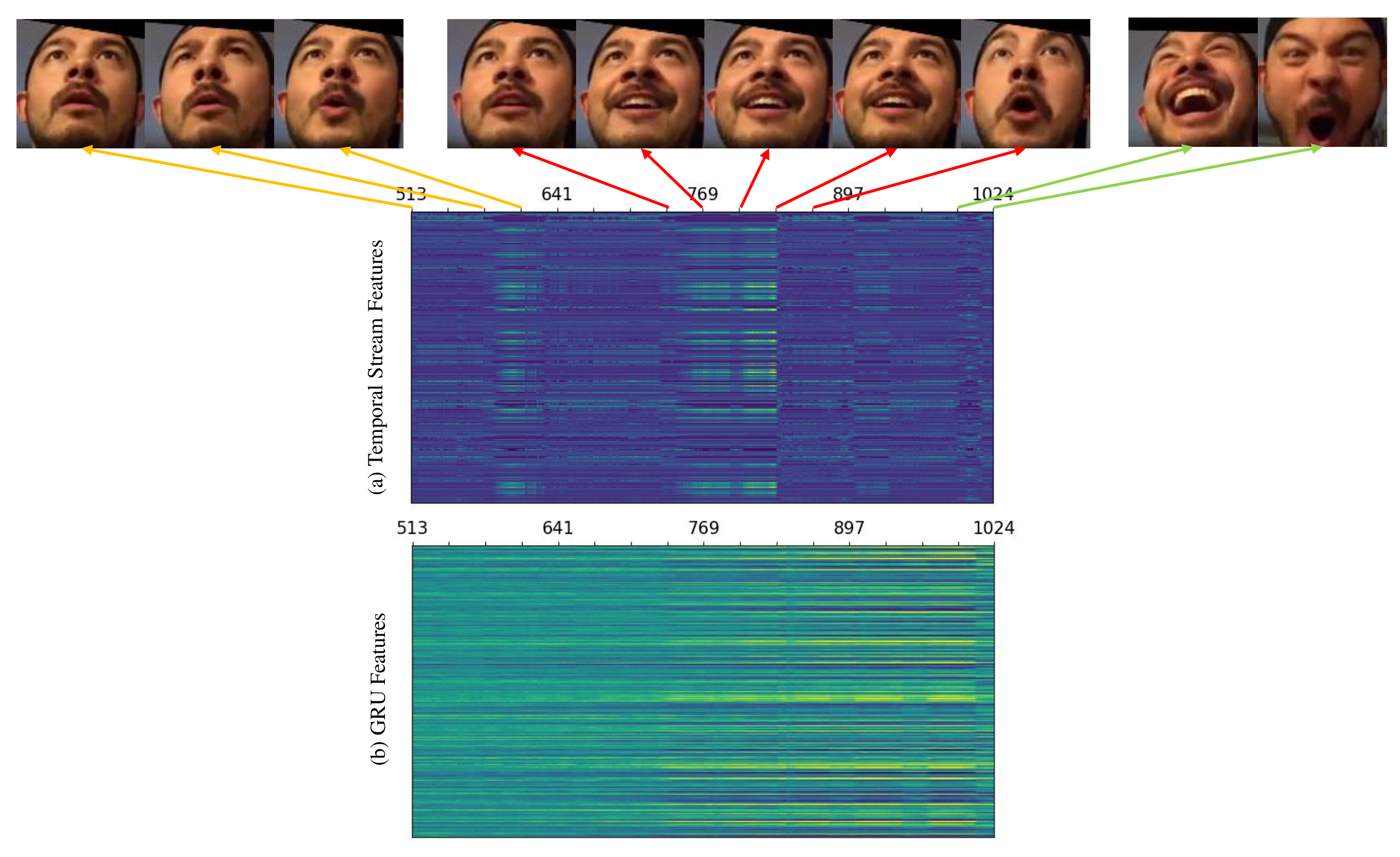} 
    } 
\caption{\textbf{Visualization of learned features}: (a) shows the 256-dimensional feature extracted by the Temporal CNN. (b) shows the 256-dimensional feature extracted by the GRU hidden units. Y axis is the index of features. X axis is the index of frames.}
\centering
\label{fig: vis}
\end{figure}

Fig.~\ref{fig: vis} visualizes the Temporal Stream CNN features (256) and the GRU hidden features (256) for a 512-frame segment. The Temporal Stream CNN features changed more dramatically, while the GRU hidden features changed more slowly. The Temporal Stream CNN features were sensitive to small and short facial movements, which is consistent with our interpretation that it extracts micro-expressions.

\subsection{Ablation Studies}
\begin{table}[b]
\small
		\begin{center}
		\resizebox{0.8\columnwidth}{!}{
			\begin{tabular}{c|c|c|c|c}
				\toprule
				 Length & 4 & 8 & 12 & 16 \\
				\hline
				 Arousal & 0.151 & 0.183 & \bf 0.212 & 0.191\\
				 Valence & 0.048 & 0.091 & \bf 0.151 & 0.092 \\
				\bottomrule 
			\end{tabular}
			}
		\end{center}
		\caption{\textbf{Varying the number of phase difference images}. Average CCC scores across 5-fold cross validation.}
		\label{tab:phase_len}
\end{table}

\subsubsection{Length of Phase Difference Images.}
When training the temporal stream on the OMG dataset, we tested the effect of varying the number of phase difference images that were fed into the CNN. The length varied from 4 to 16. Table~\ref{tab:phase_len} shows the average CCC across 5-fold cross-validation. 

When we input 12 (0.4s) phase difference images, the results were the best for both arousal and valence tasks. This optimal timescale is consistent with the timescale of micro-expressions (up to 1/3rd of a second), providing some validation for our interpretation of the temporal stream as extracting micro-expressions. Based on these results, we set the length of phase difference images to 12 for our experiments on the Aff-wild dataset.

\begin{table}
\small
    \centering
    \resizebox{\columnwidth}{!}{
        \begin{tabular}{l|cccc}\hline
        \bf Models  &\bf Arousal &\bf Valence &\bf Arousal &\bf Valence \\ \hline
        \diagbox[width=7em, height=2\line]{Temporal }{Spatial}&
          \multicolumn{2}{c}{ \textit{ResNet50} } & \multicolumn{2}{c}{ 
          \textit{None} }\\ \hline
        
        \textit{$\Delta$Phase}   & \textbf{0.377} & \textbf{0.529} & 0.193  & 0.153   \\ \hline
        \textit{Phase}   & 0.373  & 0.503 & 0.178  & 0.235 \\ \hline
        \textit{Optical Flow}   & 0.364  & 0.512 & 0.136 & 0.145 \\ \hline
         \textit{None} & 0.312   & 0.508  & - & - \\ \hline
        \end{tabular}
        }
    \caption{\textbf{Performance of different cue combinations on the OMG dataset}}
    \label{tab: omg}
\end{table}
\begin{table}
\small
    \centering
    \resizebox{\columnwidth}{!}{
        \begin{tabular}{l|cccc}\hline
        \bf Models  &\bf Arousal &\bf Valence &\bf Arousal &\bf Valence \\ \hline
        \diagbox[width=7em, height=2\line]{Temporal }{Spatial}&
          \multicolumn{2}{c}{ \textit{ResNet50} } & \multicolumn{2}{c}{ 
          \textit{None} } \\ \hline
        \textit{$\Delta $Phase}   & \textbf{0.521}   & \textbf{0.580} & 0.176  & 0.107   \\ \hline
        \textit{Phase}   & 0.516  & 0.523 & 0.118  & 0.135\\ \hline
        \textit{Optical Flow}   & 0.442  & 0.578  & 0.124  & 0.058\\ \hline
        \textit{None} & 0.411  & 0.514  & - & -\\ \hline
        \end{tabular}
        }
    \caption{\textbf{Performance of different cue combinations on the Aff-wild dataset} }
    \label{tab: aff}
\end{table}
\subsubsection{Single or Two Streams.} 
The results of the single-stream and two-stream networks on OMG and Aff-Wild are reported in Table~\ref{tab: omg} and~\ref{tab: aff}, respectively. When the temporal stream or the spatial stream is \textit{None}, the result shows the performance of a single-stream network. Otherwise, the result shows the performance of a two-stream network. We also compared the performance when using the optical flow as input to the temporal stream instead of the phase differences.

On both datasets, we found that the spatial stream was more important than the temporal stream for arousal and valence prediction. Fusing the two streams, we found large improvements on both arousal and valence prediction. This suggests the spatial and temporal streams provide complementary information.

When the temporal stream was added to the spatial stream, the improvement on arousal was greater than the improvement on valence. On the OMG dataset, the improvement on arousal was 20.8\%, while the improvement on valence was only 4\%. In Aff-Wild dataset, the improvement on arousal was 26.8\%, while the improvement on valence was only 12.8\%. Since arousal measures how active or excited the subject is, it should be more related to motion than valence, which is associated with positive or negative affectivity.

\subsubsection{Temporal Stream Inputs.} 
In order to investigate the effect of using phase differences as the motion representation, we considered two other temporal stream inputs: phase images and optical flow. We denote them by \textit{Phase} and \textit{Optical Flow} respectively. Phase difference images are denoted by \textit{$\Delta$~Phase}. When using optical flow images as input, we removed the steerable pyramid in Fig.~\ref{fig: model}, while retaining the other layers. Optical flow was computed using the off-the-shelf GPU implementation from OpenCV~\cite{brox2004high}. When the image size is 224x224, calculating the optical flow from a pair of frames takes about 0.1s, while the Complex Steerable Pyramid is 10 times faster. Table~\ref{tab: omg} and \ref{tab: aff} shows that using phase differences as input to the temporal stream outperformed the use of phase or the optical flow on both arousal and valence. 

\subsubsection{Robustness to Illumination}
Since many optical flow algorithms are based on the brightness constancy constraint, we hypothesized that the proposed phase differences are more robust than optical flow when there are illumination changes. To verify this, we randomly altered the brightness of consecutive frames in videos, creating a corrupted OMG dataset. 
We altered illumination by varying the gamma correction applied to each frame of the video. Mathematically, we applied the following transformation to each pixel in the image: $g(u) = u^{\gamma}$, where $ u\in \{0, 1\}$. When $\gamma=1$, no gamma correction is applied. When $\gamma<1$, shadows or dark regions in the image become darker. When $\gamma>1$, the image becomes lighter. We controlled the illumination variability by a parameter $\beta$ in [0, 1]. The value of $\gamma$ was sampled independently from frame to frame from a uniform distribution over [1-$\beta$, 1+$\beta$].

\begin{figure}[ht] 
\centering
\resizebox{0.8\columnwidth}!{
\includegraphics[width=\columnwidth]{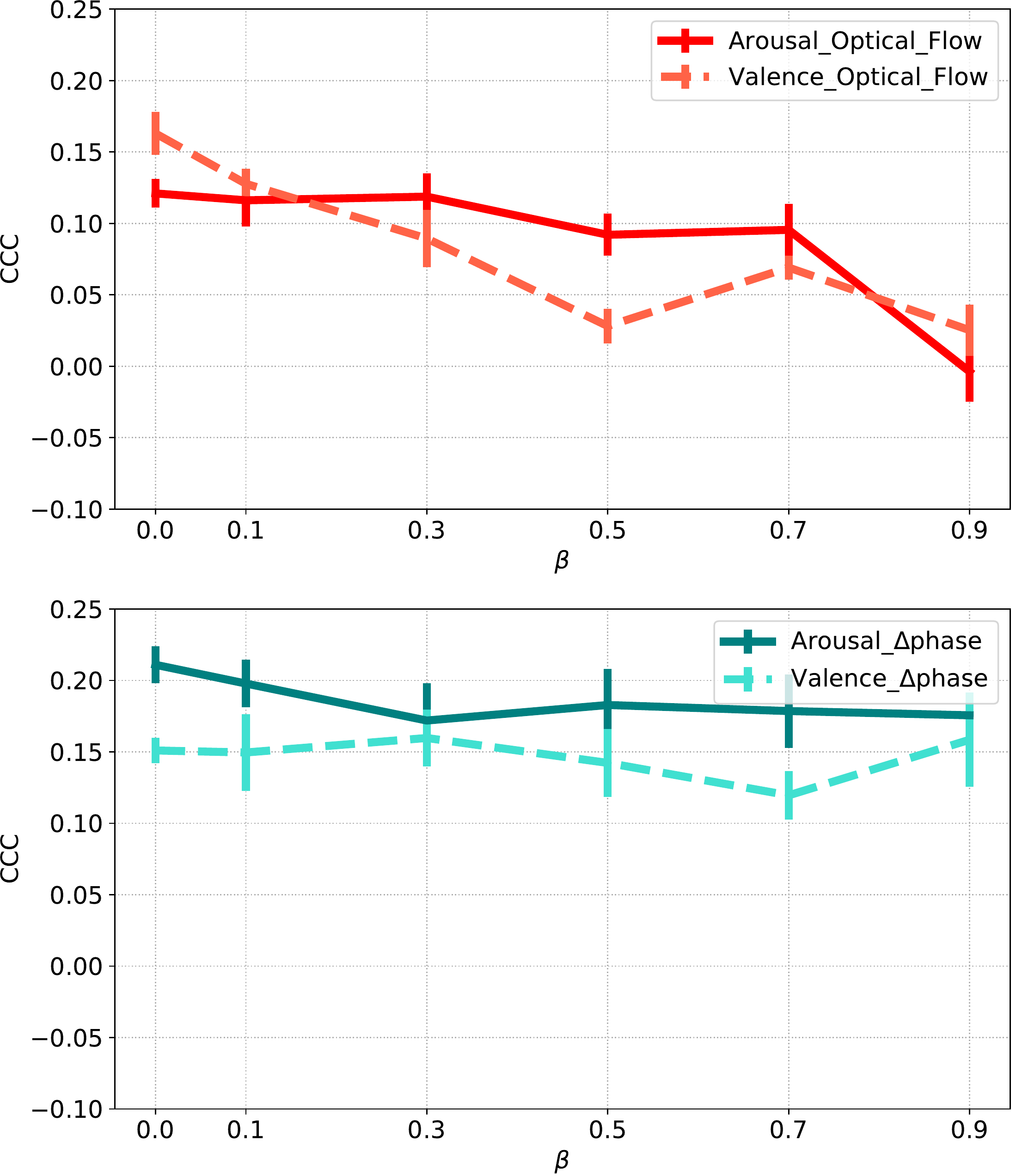} 
}
\caption{\textbf{Comparison of robustness to illumination changes}. Increasing $\beta$ leads to larger frame-to-frame variability in illumination.}
\label{fig: alter}
\end{figure}
Fig.~\ref{fig: alter} shows the experiment results. As $\beta$ became larger, the performance of the single-stream network using phase differences changed slowly but the performance of the single-stream network using optical flow dropped drastically. These results confirm that using phase differences to represent motion results is more robust under varying illumination than using the optical flow.

\section{Conclusion}
In this paper, we proposed to combine the Two-stream Network with the GRU model for video emotion recognition. We chose phase difference images as the input to the temporal stream for motion representation learning. Our method differs from the traditional Two-Stream Network which employs optical flow. To evaluate our method, we conducted experiments on two emotion datasets: the OMG dataset and the Aff-Wild dataset. From the experimental results, our model matched or exceeded the state of the art on both datasets. Phase differences, compared with optical flow, had better performance on both single-stream and two-stream results. When illumination changed, phase differences showed more robustness than optical flow, which is useful in \textit{in-the-wild} setting.

\section{ Acknowledgments}
This work was supported in part by the Hong Kong Research Grants Council under grant number 16211015 and by the Hong Kong Innovation and Technology Fund under grant number ITS/406/16FP.

\bibliographystyle{aaai}
\bibliography{aaai-2020}

\end{document}